\documentclass[final]{cvpr}

\usepackage{times}
\usepackage{epsfig}
\usepackage{graphicx}
\usepackage{amsmath}
\usepackage{amssymb}

\usepackage[pagebackref=true,breaklinks=true,colorlinks,bookmarks=false]{hyperref}

\def\cvprPaperID{772} 
\def\confYear{CVPR 2021}

\begin{document}

\title{Robust Point Cloud Registration Framework Based on Deep Graph Matching}

\author{Kexue Fu\ \ \ \ \ \ \ \ \ \ Shaolei Liu\ \ \ \ \ \ \ \ \ \ Xiaoyuan Luo\ \ \ \ \ \ \ \ \ \ Manning Wang\\
Digital Medical Research Center, School of Basic Medical Science, Fudan University\\
Shanghai Key Lab of Medical Image Computing and Computer Assisted Intervention\\
{\tt {kxfu18, mnwang}@fudan.edu.cn}
}

\maketitle

\begin{abstract}
   3D point cloud registration is a fundamental problem in computer vision and robotics. There has been extensive research in this area, but existing methods meet great challenges in situations with a large proportion of outliers and time constraints, but without good transformation initialization. Recently, a series of learning-based algorithms have been introduced and show advantages in speed. Many of them are based on correspondences between the two point clouds, so they do not rely on transformation initialization. However, these learning-based methods are sensitive to outliers, which lead to more incorrect correspondences. In this paper, we propose a novel deep graph matching-based framework for point cloud registration. Specifically, we first transform point clouds into graphs and extract deep features for each point. Then, we develop a module based on deep graph matching to calculate a soft correspondence matrix. By using graph matching, not only the local geometry of each point but also its structure and topology in a larger range are considered in establishing correspondences, so that more correct correspondences are found. We train the network with a loss directly defined on the correspondences, and in the test stage the soft correspondences are transformed into hard one-to-one correspondences so that registration can be performed by singular value decomposition. Furthermore, we introduce a transformer-based method to generate edges for graph construction, which further improves the quality of the correspondences. Extensive experiments on registering clean, noisy, partial-to-partial and unseen category point clouds show that the proposed method achieves state-of-the-art performance. The code will be made publicly available at \href{https://github.com/fukexue/RGM}{https://github.com/fukexue/RGM}. 
\end{abstract}

\section{Introduction}

\begin{figure}[t]
   \begin{center}
      \includegraphics[width=1\linewidth]{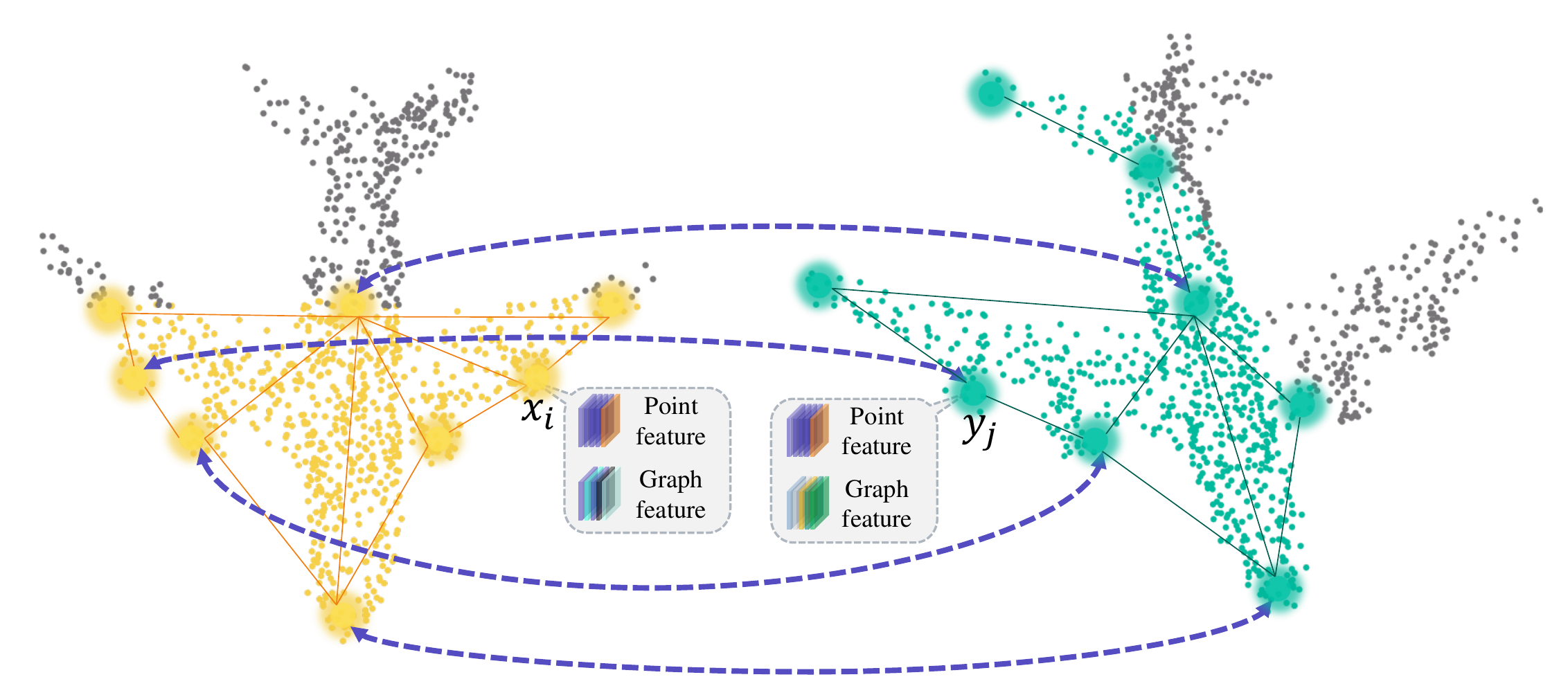}
   \end{center}
      \caption{The idea of point cloud registration based on graph matching. Dashed lines represent correspondences. Point features and graph features are the features extracted directly through points and the features extracted based on graphs, respectively. The two points $x_i$ and $y_j$ have similar point features because they have similar local geometries, but they have different graph features because the graph topologies around them are different, so they are not mismatched when graph-based matching is used.}
   \label{fig1}
\end{figure}

Rigid point cloud registration is a task that finds a rigid transformation to align two point clouds, and it has long been a fundamental task in computer vision and robotics, with many important applications, such as autopilot \cite{ref1,ref2}, surgical navigation \cite{ref3} and SLAM \cite{ref4,ref5}. There are two interlocked subproblems in point cloud registration: finding the transformation to align the two point clouds and finding the correspondences between the points \cite{ref6}. Although when the solution to one subproblem is known, the other subproblem can be easily solved, it is difficult to solve both subproblems together. Point cloud registration becomes even harder when there are outliers, which are the points with no correspondences in the other point cloud. Outliers may come from the imperfectness of the sensors used to collect the point clouds or situations in which the two point clouds to be registered are not fully overlapped.

Iterative closest point (ICP) \cite{ref7,ref8} is arguably the most widely used method for rigid point cloud registration, which starts from an initial transformation and alternately updates the correspondences and transformation. One major limitation of ICP is that it can only converge to a local optimum near the initialization, and its convergence basin is fairly small, especially when there are noise and outliers. A series of global registration methods based on branch-and-bound (BnB) \cite{ref9,ref10,ref11} have been proposed to alleviate the need for initialization by obtaining the global optimal solution, but the time-consuming BnB limits their practical applications. Another method for mitigating the need for initialization is by keypoint extraction and matching \cite{ref12,ref13}. Based on the correspondences established by matching key points, RANSAC-like methods  can be explored for registration \cite{ref12,ref13}. However, the speed and accuracy of this type of method are sensitive to outliers and repetitive geometry \cite{ref14}. Several recent methods integrate deep neural networks for establishing correspondences and a differentiable singular value decomposition (SVD) algorithm for calculating the transformation to build an end-to-end trainable network for point cloud registration, such as DCP \cite{ref15}, RPM-Net \cite{ref16} and IDAM \cite{ref17}, and they do not need transformation initialization. These methods explore deep features to establish correspondences but the discrimination ability of the features extracted from point clouds is poor, as shown in Figure~\ref{fig1}, which leads to a large proportion of incorrect correspondences and consequently devastates the registration accuracy.

In this paper, we propose a robust point cloud registration framework that utilizes deep graph matching to better handle outliers, and we denote it as RGM. By constructing graphs from point clouds to be registered and capturing  the high-order structure of the graphs, RGM can find robust and accurate point-to-point correspondences to better solve the point cloud registration problem. To the best of our knowledge, this is the first time that deep graph matching has been applied to point cloud registration. RGM contains an end-to-end deep neural network, the first part of which is a feature extractor that extracts deep local features for each point by using its neighboring points. Instead of matching these local point features directly, we construct a graph for each of the two point clouds and embed \cite{ref18} both the graph nodes (local features for each point) and graph structure (second-order or high-order structure) into node feature space. Then, we introduce an module consisting of an affinity layer, instance normalization and Sinkhorn to predict soft correspondences from the node features of the two graphs, and we denote it as AIS module. By using graph matching in the AIS module, not only the local geometry of each node but also its structure and topology in a larger range are considered in establishing correspondences so that more correct correspondences are found. In training, the binary cross-entropy loss between the predicted soft correspondences and the ground-truth correspondences are adopted, which directly promotes the network to learn better point-to-point correspondences. In testing, we use the linear assignment problem (LAP) solver \cite{ref52} based on the Hungarian algorithm \cite{ref40} to transform soft correspondences into one-to-one hard correspondences, and then SVD is employed to calculate the transformation from the hard correspondences. Similar to existing methods such as RPM-Net and ICP, we iteratively optimize the registration results.

Our main contributions are as follows:
\begin{itemize}
   \item We propose using deep graph matching to solve the point cloud registration problem for the first time. Instead of only using the features of each point, graph matching can leverage the features of other nodes and the structural information of graphs when establishing correspondences so that it can better address the problem of outliers.
   \item We introduce the AIS module to establish reliable correspondences between nodes of two given graphs. The AIS module calculates an affinity matrix between any two nodes based on the embedded features, and by analyzing the affinity matrix globally and utilizing the Sinkhorn algorithm, it can effectively reduce the proportion of incorrect correspondences.
   \item We propose using a transformer to generate soft graph edges. In registering partial-to-partial point clouds, better correspondences can be established for the overlapping parts by utilizing the attention and co-attention mechanism in the transformer.
   \item Our method achieves state-of-the-art performance on clean, noisy, partial-to-partial datasets and unseen categories datasets.
\end{itemize}

\section{Related Work}
\subsection{Traditional Registration Method}
A large proportion of traditional methods need an initial transformation and find a locally optimal solution near the initialization, in which ICP \cite{ref7,ref8} is an early and representative method. ICP starts with an initial transformation and iteratively alternates between solving two trivial subproblems: finding the closest points as correspondences under current transformation and computing optimal transformation by SVD from found correspondences. Many variants have been proposed to improve ICP \cite{ref19,ref20,ref21}. Nevertheless, ICP and its variants can only converge to a local optimum, and their success heavily relies on a good initialization. To improve the robustness to noise and outliers and enlarge the convergence basin, some methods transform point clouds into probability distributions and reformulate point cloud registration as matching two probability distributions, such as GMM \cite{ref22} and HGMR \cite{ref23}. These methods do not need to alternately solve correspondences and transformation, but their objective functions are nonconvex, so they still need a good initialization to avoid converging to a bad local optimum. Recently, a series of globally optimal methods based on BnB have been proposed, such as Go-ICP \cite{ref9}, GOGMA \cite{ref10}, GOSMA \cite{ref11}, and GoTS \cite{ref24},but they are very slow and only practical in some limited scenarios. Another line of work avoids transformation initialization by establishing correspondences. They usually first extract keypoints from the original point clouds and construct feature descriptors for them and then establish potential correspondences through feature matching \cite{ref12,ref13}. After that, RANSAC-like algorithms can be used to find the correct correspondences for registration. Different from RANSAC-like methods, FGR \cite{ref25} optimizes a correspondence-based objective function by a graduated nonconvex strategy and achieves state-of-the-art performance in correspondence-based point cloud registration. However, correspondence-based methods are sensitive to duplicate structures and partial-to-partial point clouds because a large proportion of the potential correspondences will be incorrect in these scenarios. Specifically, the lack of good initialization, a large proportion of outliers and time constraints are still big challenges for traditional point cloud registration methods.

\subsection{Learning-based Registration Method}
The developments of deep learning on point clouds allow researchers to make good use of existing research, such as PointNet \cite{ref26}, and DGCNN \cite{ref27}, to extract point cloud features for downstream tasks. These studies have stimulated the interest of using deep learning in point cloud registration. One of the earliest works is PointNetLK \cite{ref29}, which calculates global feature descriptors of the two point clouds through PointNet and iteratively uses the IC-LK algorithm \cite{ref30,ref31} to minimize the distance between the two global feature descriptors to achieve registration. PCRNet \cite{ref14} replaces the IC-LK algorithm in PointNetLK with a deep neural network. DCP \cite{ref15} utilizes transformer \cite{ref42,ref43} to compute soft correspondences between two point clouds and utilizes a differentiable SVD algorithm to calculate the transformation. Although these methods have the advantages of being fast and some of them do not need transformation initialization, they cannot effectively handle partial-to-partial point cloud registration. PRNet \cite{ref34} proposes a keypoint detector and uses the keypoint-to-keypoint correspondences in a self-supervised way to solve the partial-to-partial point cloud registration. DeepGMR \cite{ref35} extracts pose-invariant correspondences between raw point clouds and Gaussian mixture model (GMM) parameters, and then recovers the transformation from the matched Gaussian mixture models. IDAM \cite{ref17} integrates the iterative distance-aware similarity convolution module into the matching process, which can overcome the shortcomings of using inner products to obtain pointwise similarity. RPM-Net \cite{ref16} proposes a network to predict optimal annealing parameters and uses annealing and Sinkhorn \cite{ref36} to obtain soft correspondences from local features. Soft correspondences can increase robustness, but they lead to the decrease of registration accuracy, which is shown in our clean experiment. Although these methods can handle partial-to-partial point cloud registration to some extent, there is still room for improvement in their accuracy and robustness. The difference between our method and the existing learning-based methods is that we construct graphs from the original point clouds and merge structural information of the graphs into node features so that the nodes can be better matched.

Graph matching has been widely studied in computer vision and pattern recognition \cite{ref47,ref50,ref51}. Recently, learning-based graph matching has attracted considerable research interest \cite{ref37,ref38,ref18}, but, to the best of our knowledge, there is no research on using learning-based graph matching to solve the point cloud registration problem.

\begin{figure*}[t]
   \centering
   \includegraphics[width=1\linewidth]{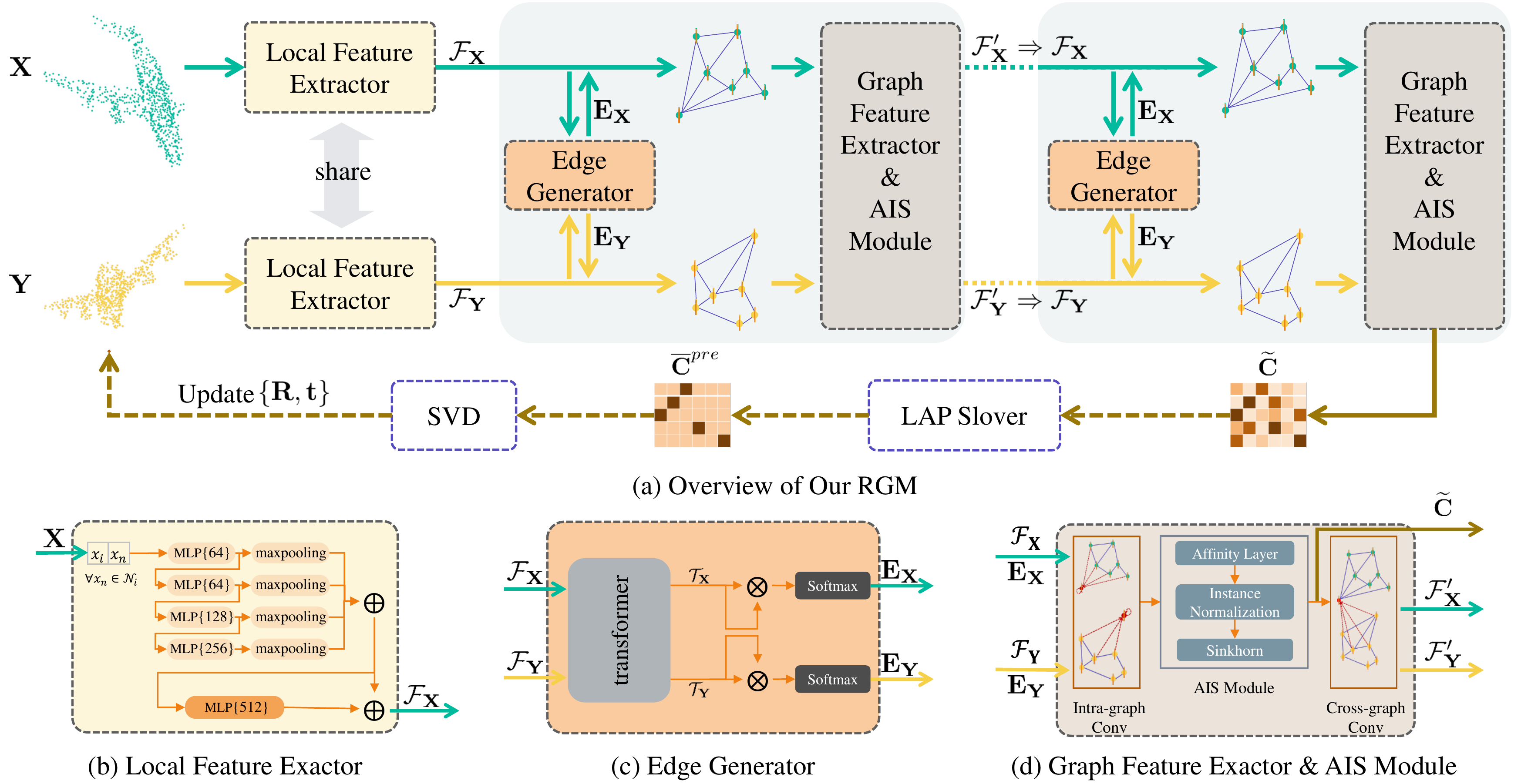}
   \caption{The pipeline of the proposed 3D rigid point cloud registration framework, RGM, where $\bigoplus$ represents concatenate features and $\bigotimes$ represents matrix multiplication. The solid lines are the data flow of both training and testing, and the dotted lines are the data flow that exists only in testing.}
   \label{fig2}%
\end{figure*}

\section{Problem Formulation}
3D rigid point cloud registration refers to estimating a rigid transformation $\left\{\mathbf{R},\mathbf{t}\right\}$ to align a source point cloud $\mathbf{X}=\left\{x_i\in\mathbf{R}^3|i=1,\cdots,N\right\}$ and a target point cloud $\mathbf{Y}=\left\{y_j\in\mathbf{R}^3|j=1,\cdots,M\right\},$ where $\mathbf{R}\in\mathbf{SO}(3)$, $\mathbf{t}\in\mathbf{R}^3$. $N$ and $M$ represent the number of points in $\mathbf{X}$ and $\mathbf{Y}$, respectively. The correspondences between points in $\mathbf{X}$ and $\mathbf{Y}$ are represented by matrix $\mathbf{C}=\left\{0,1\right\}^{N\times M}$. If $x_i$ and $y_j$ are a pair of corresponding points, $\mathbf{C}_{i,j}$ is 1; otherwise, it is 0. We first consider the simple case where there are strict one-to-one correspondences between points in $\mathbf{X}$ and $\mathbf{Y}$, in which, $N=M$. The rigid point cloud registration problem can be formulated as minimizing the following objective function:
\begin{equation}
   \boldsymbol{e}(\mathbf{C}, \mathbf{R}, \mathbf{t})=\sum_{i}^{N} \sum_{j}^{M} \mathbf{C}_{i, j}\left\|\mathbf{R} x_{i}+\mathbf{t}-y_{j}\right\|_{2}^{2},
\end{equation}
$\text {subject to } \sum_{j}^{M} \mathbf{C}_{i, j}=1, \forall i,\ \sum_{i}^{N} \mathbf{C}_{i, j}=1, \forall j,\ \mathbf{C}_{i, j} \in\{0, 1\}^{N \times M}, \forall i, j$. In the more difficult case where there are no one-to-one correspondences, the equality constraints no longer hold, and they become inequality constraints. We can introduce slack variables in $\mathbf{C}$ as in \cite{ref16} to convert inequality constraints back into equality constraints. The row constraints are converted as follows, and the column constraints are similarly converted:
\begin{equation}
   \sum_{j}^{M} \mathbf{C}_{i, j} \leq 1, \forall i \rightarrow \sum_{j}^{M+1} \mathbf{C}_{i, j}=1, \forall i \leq N.
\end{equation}
Please note that $\mathbf{C}$ becomes a $(N+1)\times(M+1)$ matrix after introducing one row and one column slack variables, and the sums of the added row and column are not restricted to be one.

In this paper, we use an end-to-end neural network to predict $\mathbf{C}$. Once we know the correspondences, the rigid transformation can be obtained by SVD.

\section{RGM}
Figure~\ref{fig2} (a) shows the overall pipeline of RGM. RGM consists of four components: local feature extractor, edge generator, graph feature extractor $\&$ AIS module and LAP-SVD. During training, we use the shared local feature extractor to extract local features for each point in $\mathbf{X}$ and $\mathbf{Y}$, and take these local features as the node features $\mathcal{F}$ of the initial graph. Next, the edge generator generates edges and builds the source graph and target graph, and the graphs are inputted into the graph feature extractor, which processes the two graphs and outputs new node features $\mathcal{F}^\prime$ and uses them to update $\mathcal{F}$. The AIS module predicts the soft correspondence matrix $\widetilde{\mathbf{C}}$ between nodes of the two graphs. By using blocks composed of three modules, the edge generator, graph feature extractor and AIS module, with the same structure but different weights $L$ times, we can obtain node features $\mathcal{F}$ with better discrimination capability and a more accurate soft correspondence matrix $\widetilde{\mathbf{C}}$. Finally, the training loss is the cross-entropy between $\widetilde{\mathbf{C}}$ and the ground truth correspondences. During test, two point clouds are first inputted into the network to obtain the soft correspondence matrix $\widetilde{\mathbf{C}}$. Then, the soft correspondences are converted to hard correspondences using the LAP solver, and the transformation is solved by SVD. We also update the transformation iteratively, similar to ICP. The details of each component are explained in the following subsections.

\subsection{Local Feature Extractor}
To establish the correspondence matrix between two point clouds, it is necessary to embed the source point cloud $\mathbf{X}$ and the target point cloud $\mathbf{Y}$ into a common feature space. We only use the coordinates of the points to build a low-dimensional local feature descriptor $\mathcal{P}$ for each point. The local feature descriptor $\mathcal{P}_{x_i}$ of $x_i$ is:
\begin{equation}
   \mathcal{P}_{x_{i}}=\left\{\left(x_{i}, x_{n}\right) \mid \forall x_{n} \in \mathcal{K}_{i}\right\},
\end{equation}
where, $\mathcal{K}_i$ represents the $K$-nearest neighboring points of $x_i$.

Low-dimensional local feature descriptors are mapped to high-dimensional local feature spaces through nonlinear functions $f_\theta$:$\mathbf{R}^{K\times6}\rightarrow\mathbf{R}^V$, where $V$ is the dimensionality of the final high-dimensional local feature. The implementation of $f_\theta$ is shown in Figure~\ref{fig2} (b), where $\theta$ represents the parameter of the nonlinear function, which consists of shared multilayer perceptron (MLP), maxpooling and concatenation. We use the high-dimensional local features as the node features $\mathcal{F}$ of the initial graph. The node feature $\mathcal{F}_{x_i}$ of $x_i$ can be expressed as follows:
\begin{equation}
   \mathcal{F}_{x_{i}}=f_{\theta}\left(\mathcal{P}_{x_{i}}\right).
\end{equation}
Inspired by the idea of the Siamese network \cite{ref41}, the two point clouds share the same local feature exactor. When the two point clouds become closer, the local features also become similar, so this structure is suitable for iterative registration.

If only the local features are used to predict the correspondences between point clouds, it is easy to obtain incorrect correspondences, especially when there are outliers. The reason is that the local features do not contain the structural information of the point cloud on a larger scale (self-correlation) and the association between the two point clouds (cross-correlation). Inspired by Wang’s research on deep graph matching \cite{ref18}, we construct graphs from point clouds and use deep graph matching to establish better correspondences. Section \ref{secEdge} describes how to build graphs from point clouds, and section \ref{secGraph} introduces how to predict the correspondences by using deep graph matching and the AIS module.

\subsection{Edge Generator Based on Transformer}\label{secEdge}
The graphs built from $\mathbf{X}$ and $\mathbf{Y}$ are denoted as source graph $\mathcal{G}_s=\left\{\mathbf{X},\mathbf{E}_\mathbf{X}\right\}$ and target graph $\mathcal{G}_t=\left\{\mathbf{Y},\mathbf{E}_\mathbf{Y}\right\}$, respectively. The graph nodes are the original points, and the graph edges are represented by the adjacency matrix $\mathbf{E}$. The node features of $\mathcal{G}_s$ and $\mathcal{G}_t$ are denoted by $\mathcal{F}_{x_i}$ and $\mathcal{F}_{y_j}$, respectively. There are trivial methods to generate the edges, such as full connection, nearest neighbor connection and Delaunay triangulation but the features of graphs cannot be effectively aggregated, as shown in Figure~\ref{fig5} (d). Inspired by the success of BERT \cite{ref42} in NLP, we introduce a transformer \cite{ref43} module to dynamically learn the soft edges of any two nodes within a point cloud. The transformer-based edge generator is illustrated in Figure~\ref{fig2} (c). The transformer consists of several stacked encoder-decoder layers. The encoder uses a self-attention layer and shared MLP to encode node features, and the decoder associates and encodes features based on the co-attention mechanism. The transformer takes node features $\mathcal{F}_\mathbf{X},\mathcal{F}_\mathbf{Y}$ as input and encodes them into embedding features $\mathcal{T}_\mathbf{X},\mathcal{T}_\mathbf{Y}$. Soft edge adjacency matrices are obtained by applying a softmax function on the inner product of the embedding features as follows:
\begin{equation}
   \mathcal{T}_{\mathbf{X}}, \mathcal{T}_{\mathbf{Y}}=f_{\text {transformer}}(\mathcal{F}_{\mathbf{X}}, \mathcal{F}_{\mathbf{Y}}),
\end{equation}
\begin{equation}
   \mathbf{E}_{\mathbf{X}}=\operatorname{softmax}(\langle\left(\mathcal{T}_{\mathbf{X}}\right)^{T}, \mathcal{T}_{\mathbf{X}}\rangle),
\end{equation}
\begin{equation}
   \mathbf{E}_{\mathbf{Y}}=\operatorname{softmax}(\langle\left(\mathcal{T}_{\mathbf{Y}}\right)^{T}, \mathcal{T}_{\mathbf{Y}}\rangle).
\end{equation}
\subsection{Graph Feature Extractor and AIS Module} \label{secGraph}
This part is shown in Figure~\ref{fig2} (d), which consists of three consecutive steps as follows:
First, we use intra-graph conv to explore the self-correlation of node features, where features are aggregated from nodes along edges within each graph. The message passing scheme between nodes is the same as PCA-GM \cite{ref18}. A node self-correlation feature $\mathcal{F}_{x_i}^{corr}$ of $\mathcal{G}_s$ is computed by intra-graph convolution as follows:
\begin{equation}
   \mathcal{F}_{x_{i}}^{corr}=\sum_{j=1}^{N}\breve{\mathbf{E}}_{i, j} * f_{adj}(\mathcal{F}_{x_{j}})+f_{self}(\mathcal{F}_{x_{i}}),
\end{equation}
and likewise for $\mathcal{G}_t$. Here, $\breve{\mathbf{E}}$ is the row normalized adjacency matrix calculated from $\mathbf{E}$, and $f_{adj}$ and $f_{self}$ are message passing functions, which are implemented by fully connected layers and ReLU.

Second, the AIS module is used to calculate a soft correspondence matrix. The AIS module consists of an affinity layer, instance normalization and Sinkhorn. An affinity matrix $\mathbf{A}$ between the two graphs is computed as follows:
\begin{equation}
   \mathbf{A}_{i, j}=(\mathcal{F}_{x_{i}}^{corr})^{T} \mathbf{W}(\mathcal{F}_{y_{j}}^{corr}),
\end{equation}
where $\mathbf{W}$ is the learnable parameter in the affinity layer. If $\mathcal{F}_{x_i}^{corr}$,$\mathcal{F}_{y_j}^{corr}\in\mathbf{R}^{Q}$, then $\mathbf{W}\in\mathbf{R}^{Q\times Q}$. 

\begin{figure}[t]
   \centering
   \includegraphics[width=1\linewidth]{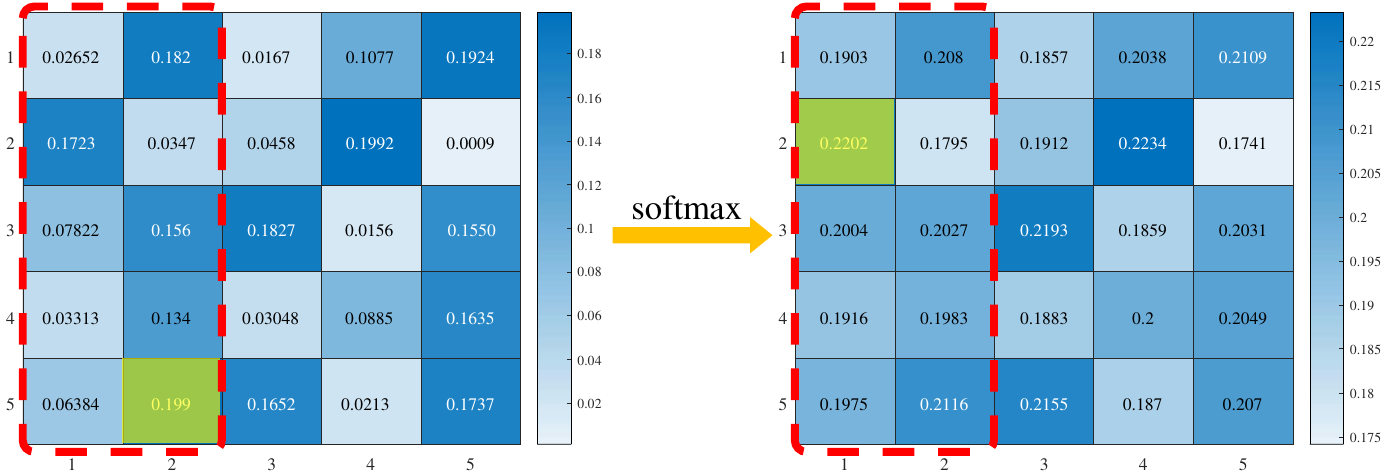}
   \caption{Columnwise softmax of the affinity matrix. Green represents the element with the largest value in the red box.}
   \label{fig3}
\end{figure}

Before using Sinkhorn to compute the soft correspondence matrix $\widetilde{\mathbf{C}}$, we need to transform $\mathbf{A}$ into a matrix with positive elements within the finite values. There are two approaches to do so, and the naïve approach is to use softmax for rows or columns. The problem with this approach is that it processes each row or column and does not consider the matrix as a whole, which may result in the problem that a smaller value in $\mathbf{A}$ is transformed into a larger value in the transformed matrix. This phenomenon is illustrated in Figure~\ref{fig3}, and we can see that the second element of the first column has a smaller value than the last element of the second column, but it becomes larger after using softmax on the two columns separately. To avoid this situation, we do not use softmax but use instance normalization \cite{ref44} to transform $\mathbf{A}$. Instance normalization considers all the elements globally and uses an exponential function to ensure that all elements are positive. For handling outliers, we add an additional row and an additional column of ones to the transformed matrix and then input it into Sinkhorn \cite{ref36} to calculate the soft correspondence matrix $\widetilde{\mathbf{C}}$ by the iterative process of alternating row and column normalizations.

Finally, we enhance the node features by exploring cross-correlation through cross-graph conv. Cross-graph conv is similar to intra-graph conv, except that features are aggregated from the node features of the other graph with edges replaced by $\widetilde{\mathbf{C}}$. The more similar the node pairs between the two graphs are, the higher the corresponding weight of $\widetilde{\mathbf{C}}$ will be. We obtain a new node feature $\mathcal{F}_{x_i}^\prime$ of node $x_{i}$ with a self-correlation feature and cross-correlation feature as follows:
\begin{equation}
   \mathcal{F}_{x_{i}}^{\prime}=f_{\text {cross}}(\mathcal{F}_{x_{i}}^{\text {corr}}, \sum_{j=1}^{\mathrm{M}} \widetilde{\mathbf{C}}_{i, j} * \mathcal{F}_{y_{j}}^{\text {corr}}),
\end{equation}
and likewise for $\mathcal{G}_t$. Here, $f_{cross}$ consists of a feature concatenate and a fully connected layer, and it is shared for $\mathcal{G}_s$ and $\mathcal{G}_t$. 

\subsection{LAP Slover and SVD}\label{secLAP}
To compute the hard correspondence matrix ${\overline{\mathbf{C}}}^{pre}$, which is binary, we sum the elements of each row and each column of $\widetilde{\mathbf{C}}$ and take out the rows and columns with a sum greater than 0.5, and apply a LAP solver based on Hungarian algorithm\cite{ref40} on the resulting matrix to obtain a binary matrix. Then, the elements of the binary matrix are assigned to a zero matrix with the shape of $\widetilde{\mathbf{C}}$ according to their position in $\widetilde{\mathbf{C}}$, and the result is the we need hard correspondence matrix ${\overline{\mathbf{C}}}^{pre}$. Finally, we take ${\overline{\mathbf{C}}}^{pre}$ as input to predict the transformation $\{\hat{\mathbf{R}},\hat{\mathbf{t}}\}$ by SVD.

\subsection{Loss}
Our loss function takes the ground truth correspondences directly as supervision, which is different from previous studies \cite{ref15,ref16,ref35} that define loss on transformation parameters. Cross-entropy loss between soft correspondence matrix $\widetilde{\mathbf{C}}$ and ground-truth correspondence matrix $\overline{\mathbf{C}}^{gt}$ is adopted to train our model. The formula is as follows:
\begin{align}
   \text{loss}=-\sum_{i}^{N}\sum_{j}^{M}(\overline{\mathbf{C}}_{i, j}^{g t} \log \widetilde{\mathbf{C}}_{i, j}+(1-\overline{\mathbf{C}}_{i, j}^{g t})\log(1-\widetilde{\mathbf{C}}_{i, j})).
\end{align}
Since our loss function is only related to the soft correspondence matrix $\widetilde{\mathbf{C}}$, the calculations in section \ref{secLAP} do not need to be differentiable.

\subsection{Implementation Details}
Our local feature extractor considers a neighborhood of $K$ = 20, and outputs final high-dimensional local features with the dimension $V$=1024. We set $L=2$ in this study. We train the network using the SGD optimizer with an initial learning rate of 1e-3. This network is implemented using PyTorch. For more details of implementation please see the supplementary material.

\section{Experiments}
\subsection{Datasets and Evaluation Metrics} \label{secDatasets}
All experiments are conducted on the ModelNet40 \cite{ref45} dataset, which includes 12,311 meshed CAD models from 40 categories. We randomly sample 2,048 points from the mesh faces and rescale points into a unit sphere. Each category consists of official train/test splits. To select models for evaluation, we take 80$\%$ and 20$\%$ of the official train split as the training set and validation set, respectively, and the official test split for testing. For each object in the dataset, we randomly sample 1,024 points as the source point cloud $\mathbf{X}$, and then apply a random transformation on $\mathbf{X}$ to obtain the target point cloud $\mathbf{Y}$ and shuffle the point order. For the transformation applied, we randomly sample three Euler angles in the range of $[0,45]^{\circ}$ for rotation and three displacements in the range of $\left[-0.5,\ 0.5\right]$ along each axis for translation. Unless otherwise noted, these settings are used by default in all experiments.

We use six evaluation metrics, and the first four are calculated from the estimated transformation parameters. They are the mean isotropic errors (MIE) of $\mathbf{R}$ and $\mathbf{t}$ proposed in RPM-Net \cite{ref16}, and the mean absolute errors (MAE) of $\mathbf{R}$ and $\mathbf{t}$ used in DCP \cite{ref15}, which are anisotropic. All rotation-related metrics are in units of degrees.

In addition, we propose a new metric, clip chamfer distance (CCD), which measures how close the two point clouds are brought to each other, and it is calculated as follows:
\begin{align}
   \operatorname{CCD}(\widehat{\mathbf{X}}, \mathbf{Y}&)=\sum\limits_{\hat{x}_{i} \in \mathbf{X}}\min(\min\limits_{y_{j} \in \mathbf{Y}}(\left\|\hat{x}_{i}-y_{j}\right\|_{2}^{2}), d)
   \notag
   \\&\ \ +\sum\limits_{y_{j} \in \mathbf{Y}}\min(\min\limits_{\hat{x}_{i} \in \mathbf{X}}(\left\|\hat{x}_{i}-y_{j}\right\|_{2}^{2}), d),
\end{align}
where $\hat{\mathbf{X}}$ is the transformed source point cloud after registration and ${\hat{x}}_i$ is the $i$th point. To avoid the influence of outliers in partial-to-partial registration, the point pair whose distance is larger than 0.1 is not included in the calculation. This is implemented by seting the threshold $d=0.1$.

Finally, we also reported the recall with MAE($\mathbf{R}$)$<1^{\circ}$ and MAE($\mathbf{t}$)$<0.1$. The best results are marked in bold font in tables.

\subsection{Comparing Methods}
We compare our method to ICP \cite{ref7}, fast global registration (FGR) \cite{ref25}, as well as three latest learning-based methods, RPM-Net \cite{ref16}, IDAM \cite{ref17} and DeepGMR \cite{ref35}. Other early learning-based methods, such as DCP and PointNetLK, are not directly compared, because experiments in \cite{ref16,ref17,ref35} have already shown that these new methods have better performance. Our method performs two iterations during the test. We adopt the ICP and FGR implemented by Intel Open3D \cite{ref46}. For IDAM and DeepGMR, we use the code provided by the authors and train the models according to the author's settings. For RPM-Net, we need to estimate the normal except in the clean experiment and use the code provided by the author. The number of iterations of RPM-Net was set to 5 according to the author's article. ICP uses the identity matrix as initialization, and none of the other methods need transformation initialization. All networks are retrained because no trained model is available.

\subsection{Clean Point Cloud}
We first evaluate the registration performance on clean point clouds and follow the sampling and transformation settings in section \ref{secDatasets}. The ground-truth correspondences are obtained by the strict correspondences between $\mathbf{X}$ and $\mathbf{Y}$. All models are trained and evaluated on clean data, and Table~\ref{tab1} shows the performance of our method and its peers. Our method achieves the best performance and greatly outperforms the strongest learning-based method. In addition, the success rate of RGM reaches 100$\%$, and most of its error metrics are close to 0, which cannot be achieved by other existing methods. Although DeepGMR also achieves a 100$\%$ success rate, its errors are larger than RGM. Some qualitative comparisons are shown in Figure~\ref{fig4} (a).
\begin{table}[t]
   \includegraphics[width=1\linewidth]{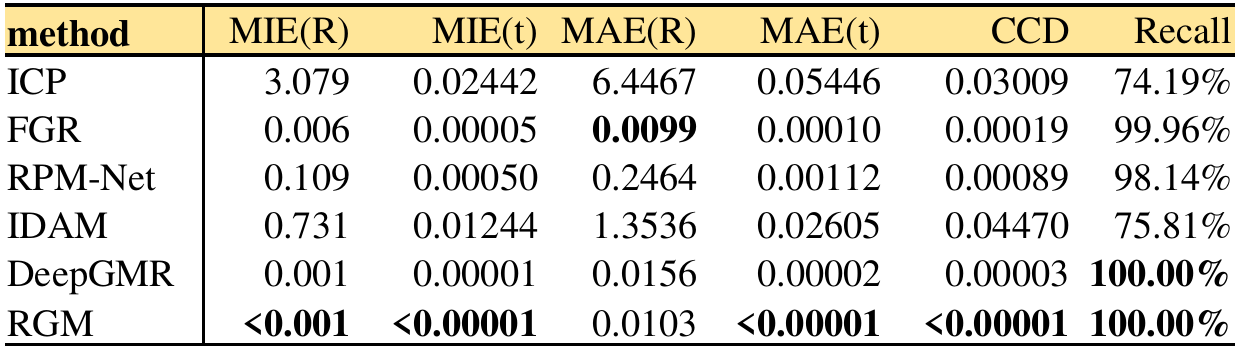}
   \caption{Performance on clean point clouds}\label{tab1}
\end{table}

\begin{table}[t]
   \includegraphics[width=1\linewidth]{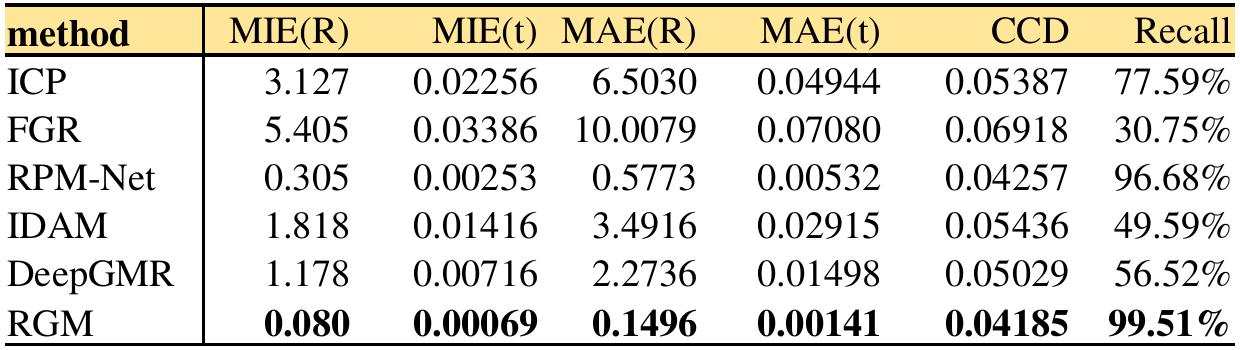}
   \caption{Performance on point clouds with Gaussian noise}\label{tab2}
\end{table}

\subsection{Gaussian Noise}
\begin{figure*}[ht]
   \centering
   \includegraphics[width=1\linewidth]{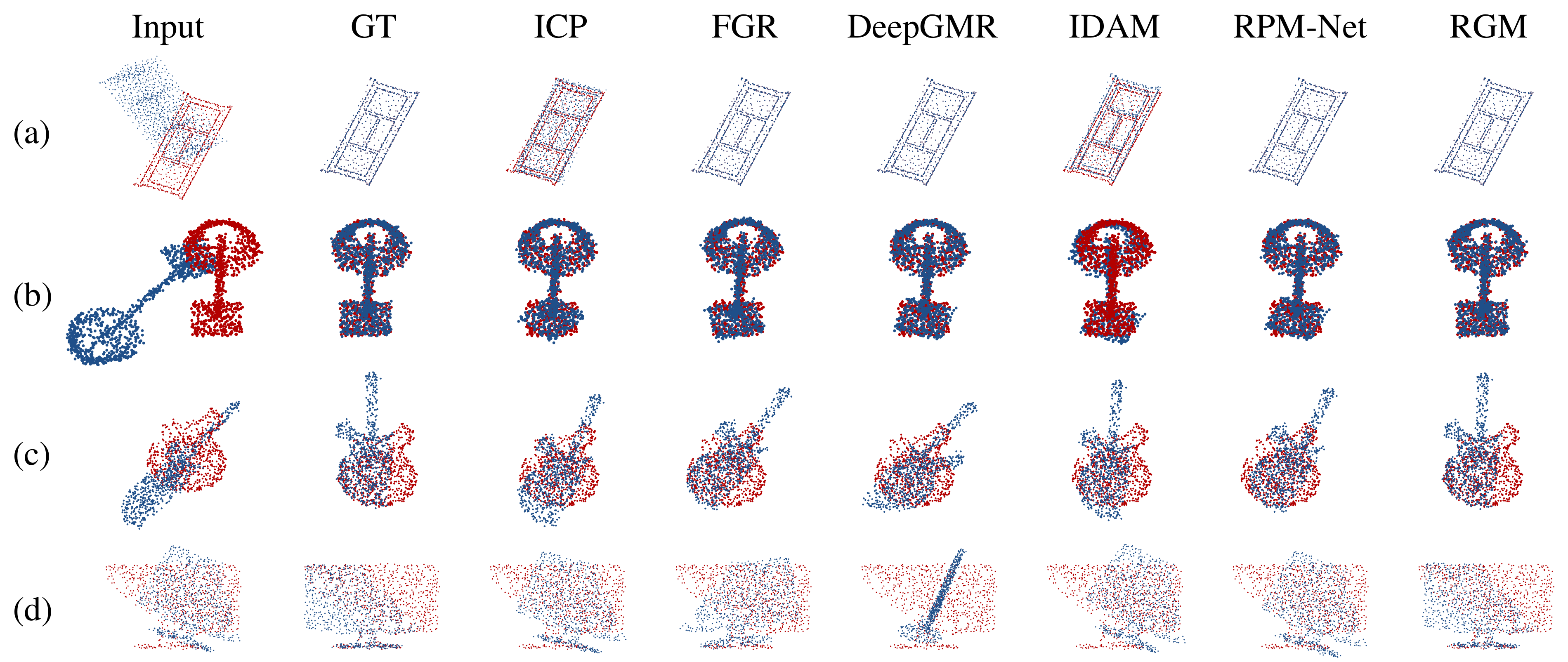}
   \caption{Qualitative registration results on ModelNet40, (a) clean, (b) noise, (c) partial-to-partial, and (d) unseen categories.}
   \label{fig4}
\end{figure*}

To evaluate the robustness to noise, Gaussian noise sampled from $\mathcal{N}\left(0,\ 0.01\right)$ and clipped to $\left[-0.05,\ 0.05\right]$ is independently added to each coordinate of the points in clean point clouds. These noises might destroy the original correspondences, so we need to rebuild them for training models that need ground truth correspondences. First, we compute the point pair distance between $\mathbf{Y}$ and $\mathbf{X}^\prime$, which is obtained by applying the ground truth transformation to $\mathbf{X}$. Then, if $x_i^\prime\in\mathbf{X}^\prime$ and $y_j\in\mathbf{Y}$ satisfy Eq.~\ref{eqmul}, they are regarded as a corresponding point pair and no longer appear in the next round calculation. Finally, we find corresponding point pairs again according to Eq.~\ref{eqmul} from the remaining points. To avoid long-distance point pairs being selected as a correspondence, we only consider the point pairs whose distance is less than 0.1. The reason why we find the corresponding point pair again from the remaining points is that the distance between the two points may not be the smallest but the second smallest, so they are not found in the first round.
\begin{equation}
   \min\limits_{x_{n}^{\prime} \in \mathbf{X}^{\prime}}(\left\|x_{n}^{\prime}-y_{j}\right\|_{2}^{2})=\left\|x_{i}^{\prime}-y_{j}\right\|_{2}^{2}=\min\limits_{y_{m} \in \mathbf{Y}}(\left\|x_{i}^{\prime}-y_{m}\right\|_{2}^{2}).\label{eqmul}
\end{equation}
All models are trained and evaluated on the noise data. The results are shown in Table~\ref{tab2}. It is obvious that our method is much more accurate than the latest learning-based methods and the traditional methods, and the recall of our method is close to 100$\%$. Some qualitative comparisons are shown in Figure~\ref{fig4} (b).

\begin{table}[t]
   \includegraphics[width=1\linewidth]{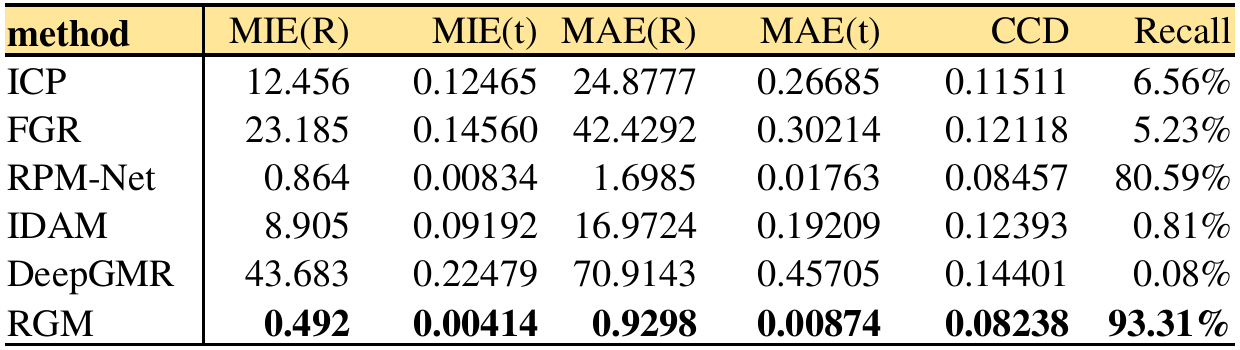}
   \caption{Performance on partial-to-partial point clouds}\label{tab3}
\end{table}

\subsection{Partial-to-Partial}

Partial-to-partial is the most challenging case for point cloud registration, and it is important because it occurs frequently in real-world applications.  To generate partial-to-partial point cloud pairs, we follow the protocol in RPM-Net \cite{ref16}, which is closer to real-world applications. For each point cloud, we create a random plane passing through the origin independently, translate it along its normal, and retain 70$\%$ of the points. All models are trained and evaluated on partial-to-partial data and the results are illustrated in Table~\ref{tab3}. Our method is obviously more accurate than the other methods, and its success rate is higher than 90$\%$. RPM-Net is the second best method, but its error is still twice as large as ours. Some qualitative comparisons are shown in Figure~\ref{fig4} (c). For the inference time of our method and the comparison methods, please refer to the supplementary material.

\begin{table}[t]
   \vspace{-0.045cm}
   \includegraphics[width=1\linewidth]{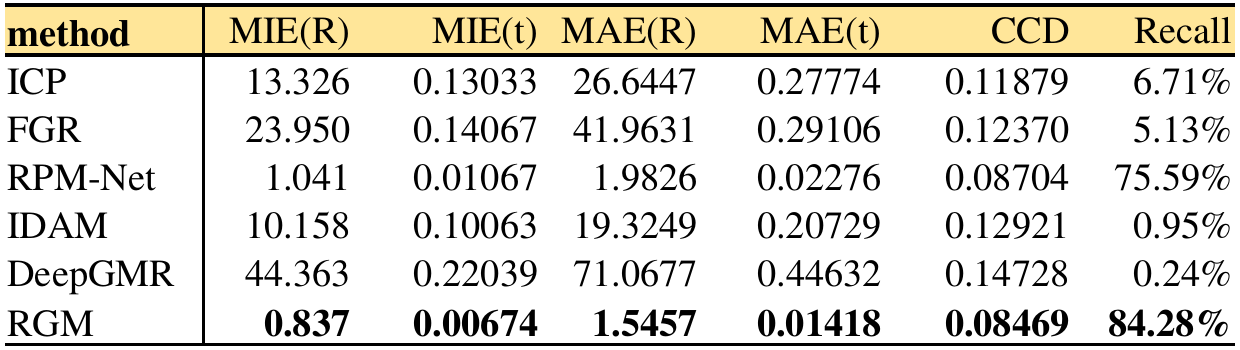}
   \caption{Performance on unseen categories point clouds}\label{tab4}
\end{table}

\subsection{Unseen Categories}

To test each method’s generalization capability on unseen shape categories, we take the official train and test splits for the first 20 categories as the training and validation sets, respectively, and test on the official test splits of the last 20 categories. Other experimental settings are the same as those in the partial-to-partial experiment. The experimental results are summarized in Table~\ref{tab4}. We find that the performance of traditional methods does not change significantly. The generalization capability of RPM-Net is also good, but it is obvious that our method works better. The other learning-based methods do not generalize well to unseen categories. Some qualitative comparisons are shown in Figure~\ref{fig4} (d).
\subsection{Ablation Studies}

\begin{figure}[t]
   \centering
   \includegraphics[width=1\linewidth]{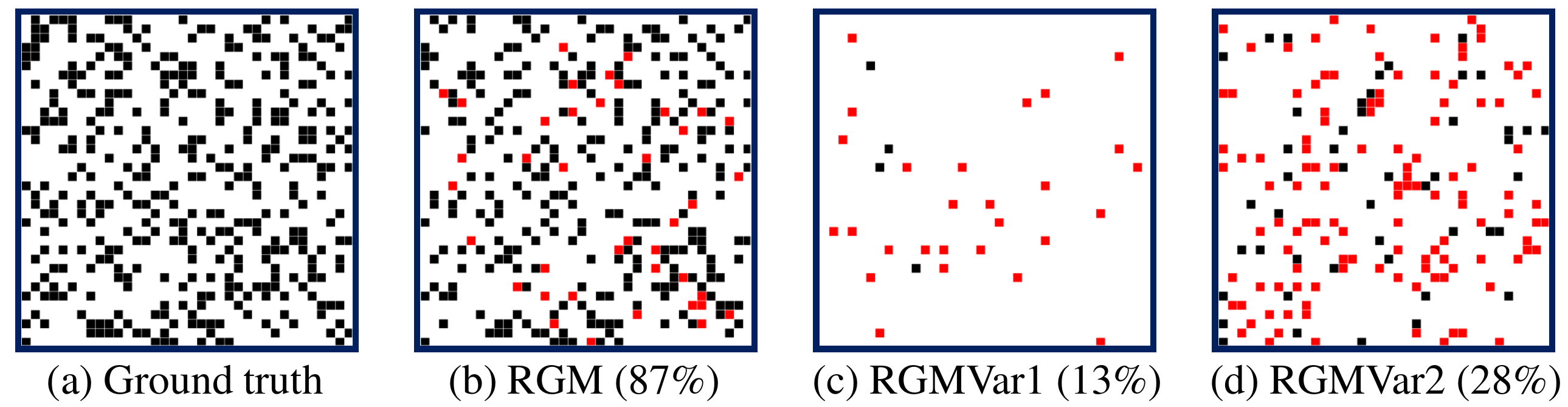}
   \caption{An illustrative case of the ground-truth correspondences and the hard correspondences generated by RGM and its variants. Black and red blocks represent the correct and incorrect correspondences, respectively. The number in brackets is the proportion of correct correspondences. Please note that there are 717 points in the two partial point clouds to be registered, and this is a sub-sampled figure with 36$\times$36 blocks. Much more correct correspondences are generated by RGM.}
   \label{fig5}
\end{figure}

In this section, we present the results of the ablation study to analyze the effectiveness of two key components. All ablation studies are performed on the partial-to-partial dataset. We analyze the two key components as follows:

To demonstrate the effectiveness of the AIS module, we design a variant to replace the AIS module, and the resulting method is denoted as RGMVar1. The variant computes the distance matrix $\mathbf{D}$ between the nodes of the two graphs by computing the L2 norm of node features, transforms $\mathbf{D}$ into a positive matrix within the finite values by the formula $e^{-\left(\mathbf{D}_{i,j}\ -\ 0.5\right)}$, and uses Sinkhorn to calculate the soft correspondences. The results are listed in the first row of Table~\ref{tab5}. We find that the registration accuracy becomes very poor by using the AIS variant, and this result shows that the proposed AIS module can effectively improve the registration performance. This is because the AIS module generates more correct matching than its variant, and an illustrative example of the hard correspondences generated by AIS and its variant is shown in Figure~\ref{fig5} (b) and (c).

\begin{table}[t]
   \includegraphics[width=1\linewidth]{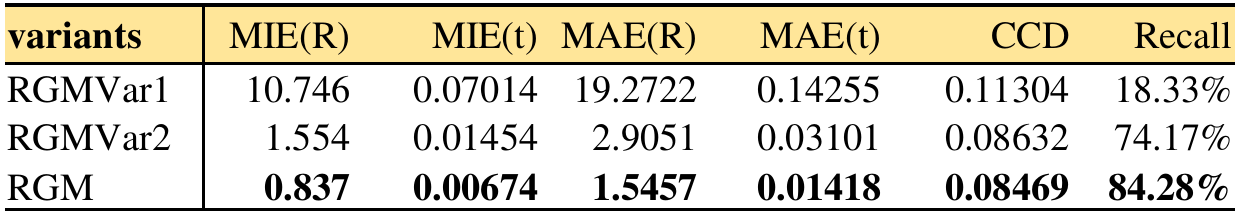}
   \caption{Ablation studies}\label{tab5}
\end{table}


To understand the importance of our edge generator, we design a variant that uses full connection edges instead of building edges by a transformer, and the resulting method is denoted as RGMVar2. The results are shown in the second row of Table~\ref{tab5}, and they are also inferior to the performance by using a transformer to generate edges. An example of the hard correspondences generated by this method is shown in Figure~\ref{fig5} (d).

\section{Conclusion}

We introduce deep graph matching to solve the point cloud registration problem for the first time and propose a novel deep learning framework RGM that achieves state-of-the-art performance. We propose the AIS module to establish accurate correspondences between the graph nodes to greatly improve registration performance. In addition, the transformer-based edge generator provides a new idea for building graph edges in addition to full connection, nearest neighbor connection and Delaunay triangulation. We think that the deep graph matching approach has the potential to be used in other registration problems, including 2D-3D registration and deformable registration. 

\newpage

{\small
\bibliographystyle{ieee_fullname}
\bibliography{egbib}
}

\end{document}